\documentclass[5pt]{article}
\usepackage[letterpaper]{geometry}
\usepackage{spconf,amsmath,epsfig}
\usepackage{enumitem}
\usepackage[english]{babel}
\usepackage{balance}
\usepackage{caption}
\usepackage{array}
\usepackage{graphicx}
\usepackage{multirow}
\usepackage{amssymb}
\usepackage{mathrsfs}
\usepackage{fancyhdr}
\begin{document}\sloppy

% Example definitions.
% --------------------
\def\x{{\mathbf x}}
\def\L{{\cal L}}

% Title.
% ------
\title{Saliency-guided video classification via adaptively weighted learning}
%
% Single address.
% ---------------
\name{Yunzhen Zhao and Yuxin Peng*\thanks{*Corresponding author.}}
\address{Institute of Computer Science and Technology, Peking University
\\ Beijing 100871, China
\\ \{zhaoyunzhen,pengyuxin\}@pku.edu.cn}
%
% For example:
% ------------
%\address{School\\
%	Department\\
%	Address\\
%   Email}
%
% Two addresses (uncomment and modify for two-address case).
% ----------------------------------------------------------
%\twoauthors
%  {A. Author-one, B. Author-two\sthanks{Thanks to XYZ agency for funding.}}
%	{School A-B\\
%	Department A-B\\
%	Address A-B}
%  {C. Author-three, D. Author-four\sthanks{The fourth author performed the work
%	while at ...}}
%	{School C-D\\
%	Department C-D\\
%	Address C-D\\
%   Email}
%

\maketitle

\begin{abstract}
Video classification is productive in many practical applications, and the recent deep learning has greatly improved its accuracy.
However, existing works often model video frames indiscriminately, but from the view of motion, video frames can be decomposed into salient and non-salient areas naturally.
Salient and non-salient areas should be modeled with different networks, for the former present both appearance and motion information, and the latter present static background information.
To address this problem, in this paper, video saliency is predicted by optical flow without supervision firstly. Then two streams of 3D CNN are trained individually for raw frames and optical flow on salient areas, and another 2D CNN is trained for raw frames on non-salient areas.
For the reason that these three streams play different roles for each class, the weights of each stream are adaptively learned for each class.
Experimental results show that saliency-guided modeling and adaptively weighted learning can reinforce each other, and we achieve the states-of-the-art results.
\end{abstract}
\section{Introduction}
%
%With the rapid growth of multimedia data, content-based video classification is productive in many practical applications, e.g. activity surveillance, video retrieval, and so on. In fact, video classification has been studied for years. Significant progress has been achieved, however there still remain great challenges. For instance, the traditional image-based feature can be used to mine the spatial information in videos \cite{1.lowe2004distinctive}, for it's a natural idea to treat frames as still images and use the appearance information to classify videos. Except for the static information, videos distinguish from images by the motion information, and some features like dense trajectories \cite{2.wang2013action} and the histograms of optical flow (HOF) \cite{17.liu2011recognizing} use motion clues to improve the effect for video classification. In the work of \cite{16.laptev2005space}, the 2D Harris corner detector is extended into 3D space for the propose of finding the interest points in videos.

With the rapid growth of video data, there is a strong need of video classification, e.g. activity surveillance, video retrieval, and so on. In fact, video classification attracts increasing research for years.
%Significant progress has been achieved, however there still remain great challenges.
Despite that significant progress has been achieved, there still remain great challenges.
The traditional image-based features can be used to mine the spatial information in videos, for it's a natural idea to treat frames as still images, and use the static information to classify videos. Some features like dense trajectories \cite{2.wang2013action}, histograms of optical flow (HOF) use motion clues to improve the performance for classification.

%In contrast to the traditional hand-crafted features, the research of deep neural networks is growing rapidly recently. Karpathy et al. \cite{7.karpathy2014large} extending the connectivity of CNN in the time domain to take advantage of local spatio-temporal information. Simonyan et al. \cite{15.simonyan2014two} propose a two-stream convolution networks for action recognition, and their architecture contains not only spatial stream ConvNet but also temporal stream ConvNet, for the propose to exploiting motion information. Ng et al. \cite{23.yue2015beyond} further propose a network combining raw frame ConvNet and optical flow ConvNet, and they use two methods to combine the two network, which are feature pooling and long short term memory (LSTM) \cite{24.hochreiter1997long}. Wu et al. \cite{8.wu2014exploring} use a deep neural network to jointly learn the feature relationships and the class relationships in videos to improve the performance, and they later propose a hybrid deep learning framework \cite{9.wu2015modeling}, which is able to model static spatial information, short-term motion, as well as long-term temporal clues together in the videos for classification.

In contrast to traditional hand-crafted features, the research of deep neural network is growing rapidly. Karpathy et al. \cite{7.karpathy2014large} extend the connectivity of CNN in the time domain to take advantage of local spatio-temporal information. Some works \cite{15.simonyan2014two}, \cite{23.yue2015beyond}, \cite{9.wu2015modeling} combine two or more networks for video classification, which jointly model static information and motion information. Long short term memory (LSTM) is often used to explore the long-term temporal clues in recent works, and 3D CNN is also a natural and suitable choice for video classification to receive such a 3-dimensional input. Ji et al. \cite{11.ji20133d} develop a 3D CNN model to extract features from both spatial and temporal dimensions, and Tran et al. \cite{10.tran2014learning} further propose a C3D feature based on 3D CNN. However, all these 3D CNN works concentrate on video frames, while ignore the information from optical flow.

In addition, these above methods use different ways to model both static and motion information \cite{15.simonyan2014two,9.wu2015modeling,10.tran2014learning} even audio information\cite{39.wu2015fusing}, but all ignore to concentrate on the division of salient and non-salient areas, and treat all the pixels of frames equally. In fact, from the view of motion, video frames often can be decompose into salient and non-salient areas, which present different information, thus should be modeled separately.
Salient areas present both appearance and motion information. Those non-salient areas usually don't present motion information, however the static background information included in these areas is beneficial for classification.
Taking football videos as an example, the salient areas contain players, footballs, or referees, and the non-salient areas mainly contain the pitch. Both of them provide helpful appearance information, but only salient areas supply the useful motion information (the moving of players and footballs).
Lazebnik et al. \cite{12.lazebnik2006beyond} use the spatial pyramid matching method in the image classification problem to pooling on different channels, and this idea is expanded into video domain in the work of \cite{13.laptev2008learning}. While the focus cues are not regularly located in the certain spatial channels of a video, which may misleads the classification models. Nguyen et al. \cite{14.nguyen2015stap} propose a spatial-temporal attention-aware pooling scheme for feature pooling. In their work, video areas with different saliency rate is distinguished, but the saliency information for training is drew by experimental instruments, and they use a lot of models, which may lead the algorithm to a high complexity. We also notice that these approaches usually take the hand-craft features, which are not robust and discriminative enough for video classification.

\begin{figure*}[!t]
  \centering
  \includegraphics[width=0.75\linewidth]{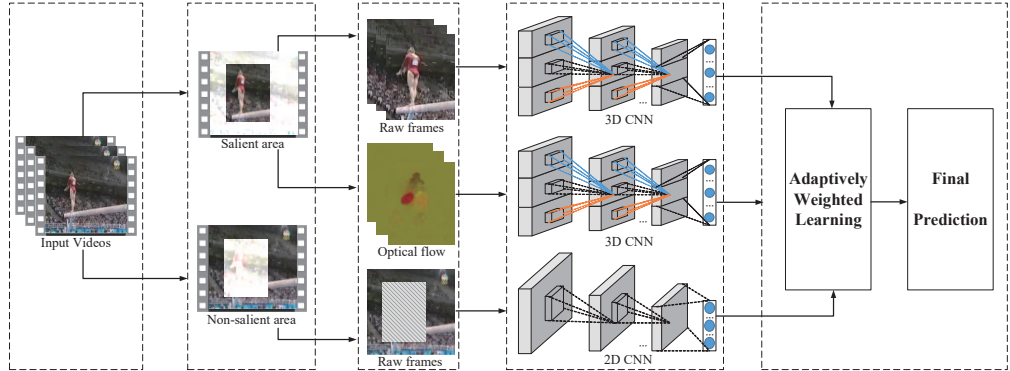}
  \caption{The framework of our proposed approach.}
  %Firstly, video saliency is predicted by optical flow without supervision. Then two streams of 3D CNN are trained individually for raw frames and optical flow on salient areas, and another 2D CNN is trained for raw frames on non-salient areas. After that, combining weights for each class are learned by adaptively weighted learning.}
  \label{fig:framework}
  \vspace{-12pt}
\end{figure*}

%In this paper, we propose a hybrid framework combining traditional 2D CNN and 3D CNN that modeling information from both salient areas and non-salient areas. Figure \ref{fig:framework} shows the framework of our proposed approach. First we eliminate the camera motion in videos and automatically predict the video saliencies by optical flow. Then we take different strategies to exploiting information from video areas of different saliency rate. For the salient areas, we train two types of 3D CNN individually to draw information from raw frames and optical flow, so as to learn the static and motion information at two aspects. For the non-salient areas, another 2D CNN is trained for the non-salient areas to draw the useful static information. The final predictions are generated by combining the outputs of the individual networks. In this way we not only model the static spatial information and motion temporal information at the same time, but use different strategies to model the salient areas and non-salient areas in videos.

In this paper, we propose a hybrid framework adaptively weighted combining 2D and 3D CNN for video classification, which not only models the static and motion information simultaneously, but also respectively models information from both salient and non-salient areas.
In detail, we first eliminate the camera motion in videos and automatically predict the video saliency by optical flow without supervision. Then we take different strategies to exploit information from salient and non-salient areas. For the salient areas, we train two types of 3D CNN individually to draw information from raw frames and optical flow, representing appearance and motion information.
For the non-salient areas, we trained a 2D CNN to model the static background information.
In the following, considering these three streams play different roles for each class, adaptive fusion weights are learned for each class specifically.

The main contributions of this work are summarized as follows:
\begin{itemize}
\item We find by estimating and eliminating the camera motion information firstly, optical flow can be used to predict the salient areas in videos without supervision. Salient areas present both appearance and motion information, and non-salient areas usually present the static information, thus should be modeled with different networks.
\item We find that 3D CNN is more suitable for capturing the motion information contained in raw frames and optical flow, and extract both appearance and motion information from raw frames and optical flow of the salient areas by 3D CNN. Both of them are proved to be useful for video classification.
\item We propose an adaptively weighted learning framework for video classification. In this way we not only model the static spatial information and motion temporal information at the same time, but also respectively model the salient and non-salient areas in videos, and finally different fusion weights for each class are learned for prediction. We also find that the saliency-guided modeling and adaptively weighted learning can reinforce each other.
\end{itemize}

The rest of this paper is organized as follows. Section 2 describes our proposed approach in details, and then in Section 3 experimental results and comparisons are discussed, followed by conclusions and future works discussed in Section 4.

\section{Methodology}
In this section, we first present our proposed hybrid framework, and then show the key components of our work.

Figure \ref{fig:framework} shows our proposed framework for video classification. From the view of motion, video data can be naturally decompose into salient and non-salient areas. Both salient and non-salient areas present the static appearance information, while the salient areas contain extra motion information, thus should be modeled separately. Inspired by the two-stream framework \cite{15.simonyan2014two}, we train three networks for video classification, which not only model the static and motion information simultaneously, but also respectively model both salient and non-salient areas.
In detail, we first segment video frames into salient and non-salient areas by the method mentioned in Section 2.1. Then we use different networks to extract information from salient and non-salient areas.
For the salient areas, we train a 3D CNN to draw the appearance and motion information from raw frames, and train another 3D CNN to draw extra motion information from optical flow of these areas. As for the non-salient areas, we concentrate on raw frames and use a traditional 2D CNN to extract the static information.
It should be noted that considering the types of videos are different, different streams play different roles for each class. For instance, sports videos usually are more sensitive to optical flow of salient areas, but for videos of birthday party, raw frames may be more important. So it's helpful to learn adaptive weights for each class to combine these three streams, which is introduced in Section 2.3.

Next, we will show the key components of the framework, which consists of prediction of the salient areas, 3D convolution and pooling, and adaptively weighted learning.

\subsection{Prediction of the salient areas}
As indicated in some works, human brains are selectively sensitive to motion. It also implies us that the movement in the videos may let us know what people pay attention to, which is useful for video research. However, in videos, the motion is caused by two reasons: the movement of the subjects in the videos, and the movement of camera. Object motion is conflated with camera motion, so if we want to find the salient areas by the information of subject motion, we should estimate and eliminate the camera motion first in videos in case of confusion.

Actually many works has discussed the estimate of camera motion for videos. Here we take the same strategy as indicated in \cite{2.wang2013action}, and the elimination of camera motion is completed by two steps.
First, we estimate the homography by finding the correspondences between two frames. Then, we use the method RANSAC to roughly correct the raw frames from the camera motion. After this step, we analysis the vectors of the trajectories in the flow field, and remove the vectors too small, which takes the same assumption indicated in \cite{2.wang2013action} that these vectors are considered to be similar enough to camera motion.

After the estimation of camera motion, we get the optical flow of videos and remove the camera motion vectors by the above strategy. For optical flow differs from frames that only motion information is kept, we can use the traditional edge detection algorithm (like Canny or Prewitt operator) to detect the location of the moving subjects in the optical flow, and get the salient areas. However there remains some troubles that we should consider:

\begin{itemize}
%\item Generally the elimination of camera motion cannot be done exact accurately, so except for the motion information of main subject which we want to attention, the background motion vectors caused by the camera motion or other reasons may be found in the corrected optical flow. However these vectors are usually small enough compared to the main subject motion vectors in optical flow, and they can be ignored by setting threshold.
\item For the reason of camera motion, we consider about the frame sequence of a certain single shot video, the location of motion subjects for each frame slightly differs. So for a single shot video, we calculate the union set for the location of the motion subjects for each frame, as the salient areas for videos.

\item For the video with multiple shots, salient areas differ in different shots. So we first segment these videos by shot, and then use above method to get the salient areas for each separate shot. Usually the salient areas are irregular and they are not convenient for the later processing, so we extend these areas to rectangular shape.
\end{itemize}

\subsection{3D convolution and pooling}
Traditional 2D convolution map the feature of two dimensions. For image classification problem, the 2D convolution is sufficient. But for the video frame sequences, 2D convolution cannot work well for the existence of temporal relationships between frames. In this way, 3D convolution can apply the feature map from three dimensions, e.g. length, width, and time sequence.

As for the formal description, 3D convolution is similar to 2D convolution. For a network, value of the unit of $j$-th feature map in the $i$-th layer is obtained by
\begin{equation}
%\label{eqn:l21norm}
    h_{ij} = f (W^k_{ij} * h_{i} + b_{ij})
\end{equation}
where $b_{ij}$ is the bias for the feature maps. For the 2D convolution, it can be further described as follow:
\begin{equation}
%\label{eqn:l21norm}
    h_{ij}^{xy} = f ( \sum_s \sum_{p=0}^{P_i-1} \sum_{q=0}^{Q_i-1} w_{ijs}^{pq} h_{(i-1)s}^{(x+p){y+q}} + b_{ij})
\end{equation}
where $P_i$ and $Q_i$ are the height and width of the convolution kernel, $s$ traverses the maps from the forward layer connected to the current feature map, and $w_{ijs}^{pq}$ is the kernel value of the position $(p, q)$ connected to the $s$-th feature map.

For the 3D convolution, the former equation expands to
\begin{equation}
%\label{eqn:l21norm}
    h_{ij}^{xyz} = f ( \sum_s \sum_{p=0}^{P_i-1} \sum_{q=0}^{Q_i-1} \sum_{r=0}^{R_i-1} w_{ijs}^{pqr} h_{(i-1)s}^{(x+p)(y+q)(z+r)} + b_{ij})
\end{equation}
we can see the specific dimension $R_i$ is what 3D convolution differs from 2D convolution, for the kernel is of 3-dimension. 2D convolution apply on images and we obtain images. 3D convolution apply on frame sequence and we obtain frame sequence. In this way, both the spatial and temporal information can be reserved. What's more, the difference between 2D and 3D pooling is just the same as between 2D and 3D convolution.

\subsection{Adaptively weighted learning}

%Next, we will introduce our adaptively weighted learning approach. As we have obtained the softmax score of each stream, we can simply combine the score of each stream and get the final results. However, due to the reason that these three streams play different roles for each class, it's better to learn different combining weights for each class.

Next, we will introduce our adaptively weighted learning approach. As we have obtained the softmax score of each stream, we can simply combine the score of each stream and get the final results. However, due to the reason that these three streams play different roles for each class, it's better to learn different combining weights for these semantic classes.

Formally, we denote the prediction score of $i$-th training data in $j$-th class as $S_i^j = [{s_{i,1}^j}^T, {s_{i,2}^j}^T, {s_{i,3}^j}^T]^T \in \mathbb{R}^{3 \times c}$, where $c$ denotes the number of semantic class, $s_{i,m}^j \in \mathbb{R}^{1 \times c}$ stands for the score of $m$-th stream for $i$-th training data in $j$-th class, and the fusion weight for $j$-th semantic class as $W_j = [w_{j,1}, w_{j,2}, w_{j,3}]$, with the restriction that $\sum_{i=1}^3 w_{j,i} = 1, w_{j,i} > 0$.
The fusion weight for each class is learned separately, and to obtain the weight $W_j$, we define the objective function as:
\begin{equation}
    arg \max \limits_{W_j} \mathcal{P}_j - \lambda \mathcal{N}_j
\label{equ:firstequ}
\end{equation}
$\mathcal{P}_j$ is defined as:
\begin{equation}
    \mathcal{P}_j = \sum_{i=1}^{n_j} W_j S_i^j J_j
\end{equation}
where $n_j$ stands for the number of training data in $j$-th class. $J_j = [0,...,1,...,0]^T \in \mathbb{R}^{c \times 1}$, with the $j$-th element being $1$, and other element being $0$. The goal of maximizing $\mathcal{P}_j$ is to maximize the product of $W_j$ and the $j$-th column vector of $S_i^j$. Similarly, we define:
\begin{equation}
    \mathcal{N}_j = \sum_{\{k=1,k \neq j\}}^c \sum_{i=1}^{n_k} W_j S_i^k J_j
\end{equation}
It means minimizing the product of $W_j$ and the $j$-th column vector of $S_i^k$ ($k \neq j$). $\mathcal{P}_j$ and $\mathcal{N}_j$ consider the relationship of positive and negative training data for $W_j$ respectively, and $\lambda$ is the parameter to balance the weight of positive and negative samples. Then the equation (4) can be transformed to
\begin{equation}
    arg \max \limits_{W_j} W_j (\sum_{i=1}^{n_j} S_i^j J_j - \lambda \sum_{\{k=1,k \neq j\}}^c \sum_{i=1}^{n_k} S_i^k J_j)
\end{equation}
with the restriction
\begin{equation}
    \sum_{i=1}^3 w_{j,i} = 1, w_{j,i} > 0
\end{equation}
and the best weights can be learned by linear programming easily.

As for the test data, we first calculate and stack the softmax score of each stream, which is denoted as $S_t = [{s_{t,1}}^T, {s_{t,2}}^T, {s_{t,3}}^T]^T \in \mathbb{R}^{3 \times c}$, and the classification is predicted by
\begin{equation}
    arg \max \limits_{i} W_i S_t J_i
\end{equation}
so different fusion weights are considered for each class specifically, and the final result are determined by the highest fusion score.

\section{Experiment}

\subsection{Datasets}
We evaluate the effectiveness of our proposed approach on two popular datasets: 1) UCF-101; 2) Columbia Consumer Videos. UCF-101 dataset is one of the most popular action recognition datasets. It consists of 13320 video clips, which are classified into 101 classes. All the videos are collected from the YouTube website. For the splitting of training and test set, we just follow the common three splits for the dataset.
Columbia Consumer Videos (CCV) is a consumer video database. The CCV database contains 9317 web videos of over 20 semantic categories, and we use the split of 4659 videos for training and 4658 videos for test.

For UCF-101 dataset, we measure the results by averaging accuracy over three splits. For CCV dataset, we use mean AP (mAP) to measure the effectiveness of our proposed framework. In detail, we first calculate the average precision (AP) for each class, then mAP is reported for the whole dataset.

\subsection{Implementation details}
We use the same strategy to train deep networks on UCF-101 and CCV datasets, except that for the reason of videos in CCV dataset often consists of multi shots, we first segment these videos by shot and then get the salient area for each shot separately.
The optical flow is obtained by the ready-made tool in OpenCV by GPU.

For 3D CNN, each video/shot is split into 16-frame clips, and the frames/optical flow resized into $128 \times 171$, so the input dimension of 3D CNN is $16 \times 128 \times 171$.
The 3D CNN architecture is generally the same as mentioned in the work \cite{10.tran2014learning}, for it gets the state-of-the-art results with 3D CNN. The proposed 3D CNN architecture constitutes of 8 convolution layers, 5 max-pooling layers, and 2 fully connected layers, as well as a followed softmax output layer. In detail, the conv1 layer has 64 output units, conv2 has 128 output units, conv3a and conv3b has 256 output units. As for conv4a, conv4b, conv5a and conv5b, the number is 512. All these convolution layers are with the 3D filters of the kernel size $3 \times 3 \times 3$ and stride $1 \times 1 \times 1$. As for pooling layer, pool1 layer is with the kernel size $1 \times 2 \times 2$ and stride $1 \times 2 \times 2$, and other pooling layers e.g. pool2, pool3, pool4, pool5 is with the kernel size $2\times 2 \times 2$ and stride $2 \times 2 \times 2$. For each fully connect layer (fc6 and fc7), the number of output units is 4096.
The training is done with the mini-batch size of 30 examples by stochastic gradient descent.
%As for learning rate, we first set the number to $3 \times 10^{-3}$ and it decreases to $3 \times 10^{-4}$ after 100K iterations, then decrease to $3 \times 10^{-5}$ after 200K iterations. The same settings are used for 3D CNN on optical flow except for the learning rate, which is $10^{-2}$ initially and decreases to $10^{-3}$ after 100K iterations, then decreases to 1e-4 after 200K iterations.
For 2D CNN on raw frames, we use VGG\_19 network to extract static information, for it¡¯s good performance on image classification. It is first pre-trained with the ILSVRC-2012 dataset and then fine-tuned by the video data, which is similar to \cite{9.wu2015modeling}.
%The learning rate starts from $10^{-3}$ and after 14K iterations decreases to $10^{-4}$, then to $10^{-5}$ after 20K iterations.
For each separate stream, the predict score of a video is obtained by averaging all it's clip results. For adaptively weighted learning, we set the parameter $\lambda$ as $5 \times 10^{-3}$.

\subsection{Experimental results}
\begin{table}[t]
\begin{center}
\scalebox{0.9}{
\begin{tabular}{|l|c|c|}
\hline
    &  UCF-101 & CCV \\
    \hline
    3D CNN on frames of salient areas & 86.2 & 77.4 \\
    3D CNN on optflow of salient areas & 88.1 & 71.9 \\
    2D CNN on frames of non-salient areas & 77.8 & 74.6 \\
    \hline
    Frame 3D + optflow 3D & 90.4 & 80.1 \\
    Frame 3D + frame 2D & 90.7 & 81.6 \\
    Optflow 3D + frame 2D & 91.2 & 78.4 \\
    \hline
    Late fusion of all three streams & {\bf 92.1} & {\bf 83.5} \\
    \hline
\end{tabular}}
\end{center}
\vspace{-12pt}
\caption{Experimental results comparing different combinations of these three streams.}
\label{table:results1}
\vspace{-12pt}
\end{table}

We show our experiment results in Table \ref{table:results1} and \ref{table:results2}. We first measure the results of each stream separately, then combine each two of streams to find out the complementarity of them by averaging their softmax scores (late fusion), and finally combine all the streams (late fusion) to show the effect of combining these three streams.
Then we compare the results of whether or not modeling saliency, and use different fusion methods to combine these three steams, to evaluate the effectiveness of modeling saliency, and the adaptively weighted learning method.

The first group of Table \ref{table:results1} compares the results of each stream. On UCF-101, we see that the result of 3D CNN on optical flow of salient areas is higher than 3D CNN on frames of salient areas, and there exists an obvious decrease from 3D CNN on frames to 2D CNN on frames of non-salient areas. It is for the reason that since UCF-101 is an action recognition dataset, how to mine the motion information is the key point for classification. 2D CNN of non-salient areas concentrate on the background, in this case it get less information than the other two stream. On CCV dataset, the condition is quite different from the UCF-101 dataset. The 3D CNN on optical flow, which performs best on the UCF-101 dataset, achieve lower result than the other two streams. It is for the reason that since videos in CCV dataset is more diversification, the motion information seems to be noisy, and the training data for classification is limited. For the reason that 3D CNN on frames of salient areas mine the appearance and motion information of salient areas at the same time, it achieves best result among all the three streams.

The second group of Table \ref{table:results1} shows the performance of combining each two of streams. On UCF-101 dataset, the results of three combination ways get roughly the same results, since the three convolution networks draw different information from different aspects, and they can complement each other.
%However combining 3D CNN on optical flow of salient areas and 2D CNN on frames of non-salient areas gets slightly better result than the other two, since it can maximizing  mine the static and motion information at the same time.
For CCV dataset, the combination of 3D CNN on optical flow and 2D CNN on frames doesn't work very well, and 3D CNN on frames get better result with the help of 2D CNN on frames.
The last line of Table \ref{table:results1} shows the results combining all the three streams. We can see a 0.9\% improvement on UCF-101 dataset and a 1.9\% improvement on CCV dataset, comparing to the highest two-stream results. So each mentioned stream plays an important role in the proposed framework, and they can incorporate each other to get better performance.

%\begin{table}[t]
%\begin{center}
%\scalebox{0.9}{
%\begin{tabular}{|c|c|c|}
%\hline
%    UCF-101 & average & adaptively weighted \\
%    \hline
%    frame 3D + optflow 3D & 90.4 & 91.6 \\
%    frame 3D + frame 2D & 91.7 & 92.4 \\
%    optflow 3D + frame 2D & 91.5 & 92.8 \\
%    \hline
%    all three streams & 92.7 & {\bf 93.9} \\
%    \hline
%    \hline
%    CCV & average & adaptively weighted \\
%    \hline
%    frame 3D + optflow 3D & 80.1 & 81.7\\
%    frame 3D + frame 2D & 81.6 & 83.1\\
%    optflow 3D + frame 2D & 78.3 & 79.8\\
%    \hline
%    all three streams & 84.5 & {\bf 85.4}\\
%\hline
%\end{tabular}}
%\end{center}
%\vspace{-12pt}
%\caption{The experimental results comparing different fusion approach.}
%\label{table:results2}
%\end{table}

\begin{table}[t]
\begin{center}
\scalebox{0.85}{
\renewcommand{\multirowsetup}{\centering}
\begin{tabular}{|c|c|c|}
\hline
     {UCF-101} & without modeling saliency & {modeling saliency} \\
     \hline
     Late fusion & 91.2 & 92.1 \\
     Early fusion & 91.3 & 92.3\\
     Weighted fusion & 91.8 & 92.5\\
     \hline
     \textbf{Ours} & {92.8} & {\textbf{94.4}} \\
     %{\bf adaptively weighted learning} & 92.8 & {\bf 94.4}\\
     \hline
     \hline
     {CCV} & without modeling saliency & {modeling saliency} \\
     \hline
     Late fusion & 82.7 & 83.5 \\
     Early fusion & 82.9 & 83.6\\
     Weighted fusion & 83.2 & 83.9\\
     \hline
     \textbf{Ours} & {84.4} & {\textbf{85.9}} \\
     %{\bf adaptively weighted learning} & 84.4 & {\bf 85.9}\\
\hline
\end{tabular}}
\end{center}
\vspace{-12pt}
\caption{Experiments comparing the result of whether or not modeling saliency, and different fusion approaches.}
\label{table:results2}
\vspace{-12pt}
\end{table}

Table \ref{table:results2} compares the results of whether or not modeling saliency, and the results of different fusion approaches.
To validate the effectiveness of modeling saliency, we compare the results of our approach, and the results that without dividing video into salient and non-salient areas, but fusing the streams of 3D CNN on whole frames, 3D CNN of optical flow, and 2D CNN on whole frame directly. From Table \ref{table:results2} we see that experimental results are promoted with modeling saliency, no matter which fusion strategy we choose. It is for the reason that with modeling saliency, we can specifically select the models appropriate for different streams, which is beneficial.

As for combining multi-streams, we compare the results of (1) averaging the softmax scores of each stream (late fusion), (2) fusing features obtained by each stream (the output of FC7 layer) and training a SVM for classification (early fusion), (3) fusing the scores of each stream with weights obtained by cross-validation \cite{15.simonyan2014two}, and (4) fusing the scores by the adaptively weighted learning approach.
It is reported that the results of early fusion are slightly better than late fusion, and by weighted fusion we can obtain small increase than early fusion.
From the final results we find the fusion results with modeling saliency can be improved by adaptively weighted learning greatly, and see an approximate improvement of the final results on both datasets. It is for the reason that we not only model different areas and different information specifically, but also train adaptive weights for each class, which can reinforce each other.

\subsection{Comparison with states-of-the-arts}

\begin{table}[t]
\begin{center}
\scalebox{0.9}{
\begin{tabular}{|c|c|c|c|}
\hline
    Method & UCF-101 & Method & CCV \\
    \hline
    Simonyan et al. \cite{15.simonyan2014two} & 88.0 & Xu et al. \cite{44.xu2013feature} & 60.3 \\
    Ng. et al. \cite{23.yue2015beyond} & 88.6 & Jhuo et al. \cite{46.jhuo2014discovering} & 64.0 \\
    Zha et al. \cite{40.zha2015exploiting} & 89.6&  Ye et al. \cite{48.ye2012robust} & 64.0 \\
    Tran et al. \cite{10.tran2014learning} & 90.4 & Liu et al. \cite{42.liu2013sample} & 68.2 \\
    Wu et al. \cite{9.wu2015modeling} & 91.3 & Wu et al. \cite{9.wu2015modeling} & 83.5 \\
    Wu et al. \cite{39.wu2015fusing} & 92.2 & Wu et al. \cite{39.wu2015fusing} & 84.0 \\
    \hline
    \bf{Ours} & \bf{94.4} & \bf{Ours} & \bf{85.9} \\
    \hline
\end{tabular}}
\end{center}
\vspace{-12pt}
\caption{Experimental results comparing with states-of-the-arts.}
\label{table:results3}
\vspace{-12pt}
\end{table}

To objectively evaluate our proposed framework, several state-of-the-art methods are compared, and Table \ref{table:results3} show the results.
On both UCF-101 dataset and CCV dataset, our method achieves the best result among all the approaches, for the reason that we not only use different strategies to model the salient and non-salient areas in videos, but also learn adaptive weights for each class.
%It should be noted that in the work \cite{39.wu2015fusing} they get their results combining both visual and audio information, but we also get higher results.
It should be noted that we list the results without using the audio stream in \cite{39.wu2015fusing}. But comparing to their results with audio stream, we also gain a 1.8\% promotion on UCF-101 dataset, and a 1.0\% promotion on CCV dataset.
We also see that results on CCV datasets has an obviously increase since the work \cite{9.wu2015modeling}, which benefits from the appliance of deep learning.

\section{Conclusions}
In this paper, we propose an adaptively weighted learning framework for video classification.
%We first predict video saliencies by optical flow. Then for salient areas, we train two types of 3D CNN individually for raw frames and optical flow. Another 2D CNN is trained for the raw frames of non-salient areas. Then different weights for each class are learned to predict the semantic categories for videos.
%We not only model the static and motion information simultaneously, but also jointly model information from both salient areas and non-salient areas.
We divide video frames into different areas by saliency rate, and model the static and motion information simultaneously. In the following, the adaptive weights are learned for each class specifically.
Experiments on UCF-101 and CCV datasets show that our framework achieves better performance, and both of the streams are expected to improve video classification results.
We also find that modeling different areas and different information specifically, and training adaptive weights for each class can reinforce each other.
For the future work, we believe that with the appropriate help of manual indicating and handcrafted labeling, saliency regions can be extracted more accurately, especially for the complex videos. Also we will seek for a more effective way to complete the fusion process and get better performance.

\section{Acknowledgement}
This work was supported by National Natural Science Foundation of China under Grants 61371128 and 61532005.

\balance

\end{document}